\pdfoutput=1  
\documentclass[10pt,twocolumn,letterpaper]{article}

\usepackage[pagenumbers]{wacv}

\providecommand{\nolinenumbers}{}

\newcommand{\mtla}{MTLA}

\newcommand{\ga}{GA}
\newcommand{\auroc}{AUROC}
\newcommand{\mAP}{AP}

\definecolor{wacvblue}{rgb}{0.21,0.49,0.74}
\usepackage[pagebackref,breaklinks,colorlinks,allcolors=wacvblue]{hyperref}

\def\wacvPaperID{380}
\def\confName{WACV}
\def\confYear{2027}

\title{Propose and Attend: Training-free MLLM Grounding Confidence via Multi-Token Localized Attention}

\author{Daniel Shalam \quad Emanuel Ben Baruch \quad Avi Ben Cohen \quad Tal Remez\\
Amazon\\
{\small Code: \url{https://github.com/TalRemez/MTLA.git}}
}

\begin{document}
\maketitle
\begin{abstract}
Multimodal large language models can emit localized predictions, bounding
boxes for objects and temporal windows for video and audio events, but they
hallucinate these regions prolifically. The model's own token
log-probabilities are nearly uninformative: they conflate grounding quality
with input ambiguity, and coordinate tokens become near-deterministic once
the model commits. We propose \textbf{Multi-Token Localized Attention}
\textbf{(\mtla{})}: a training-free, post-hoc score that measures how strongly a
prediction's tokens attend to the region they claim. Prior
attention-based detectors, which sum attention over the entire input modality and
read a single response token, are weaker special cases; we show that summing only
within the claimed region and aggregating
across all prediction tokens recovers a stronger grounding signal. The same
recipe applies almost trivially to other modalities and tasks: object
detection in images and temporal localization in video and audio. Across multiple MLLM families and three modalities,
\mtla{} improves hallucination \auroc{} by $+7$ to $+21$ over the best prior
training-free baseline. Used as a confidence score for re-ranking, it nearly doubles the
zero-shot COCO detection AP of an open-source $8$B generalist (from $20.4$ to
$37.0$), narrowing the gap to supervised detectors without any task-specific
training.
\end{abstract}

\begin{figure}[t]
  \centering
  \includegraphics[width=\linewidth]{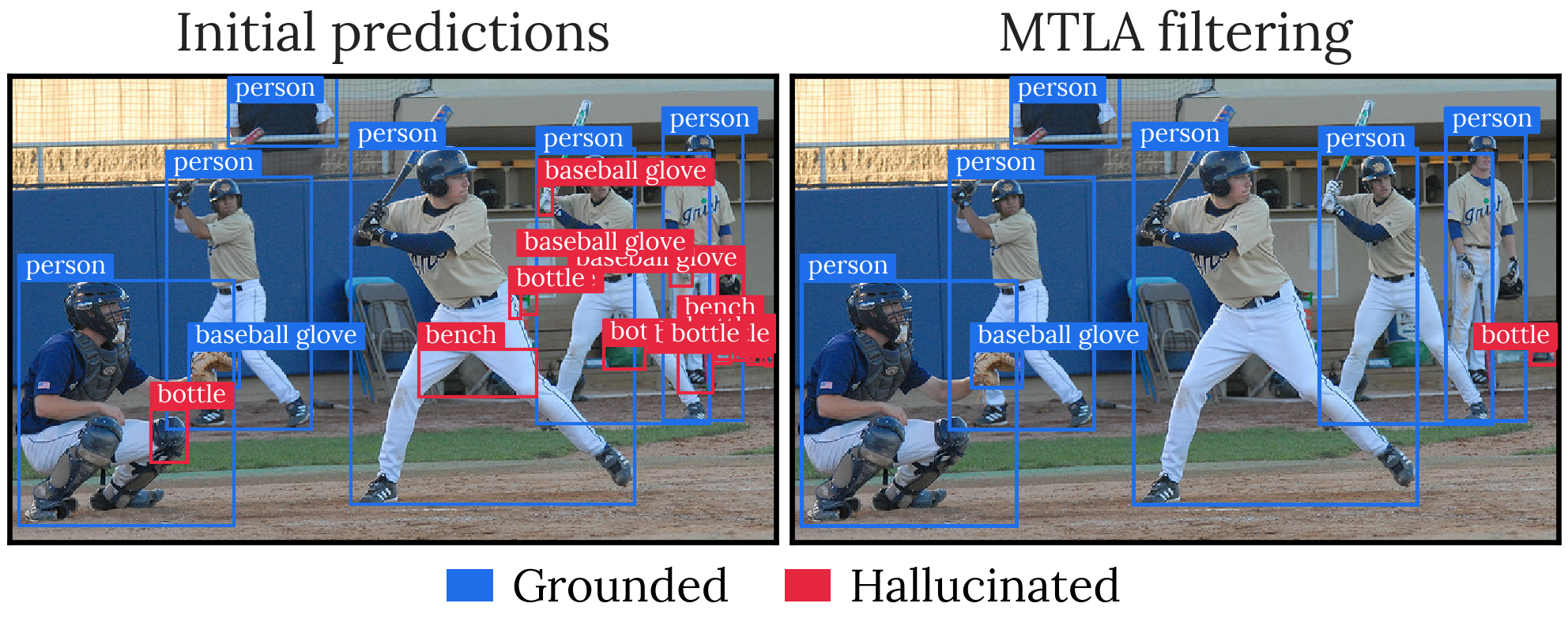}
  \\[5pt]
  \includegraphics[width=\linewidth]{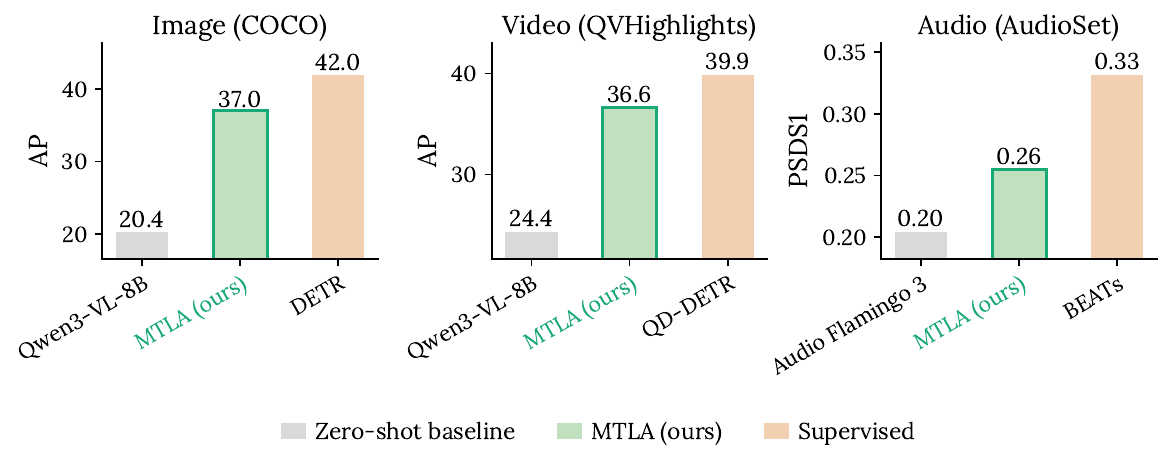}
  \setlength{\abovecaptionskip}{4pt}
  \caption{
\textbf{\mtla{} enables reliable localization from MLLMs.}
\emph{Top:} MLLM detection outputs contain many hallucinated predictions
(red); \mtla{} uses the model's own attention to estimate localization
confidence and suppress them, retaining grounded predictions (blue).
\emph{Bottom:} applied post hoc, \mtla{} ($N{=}16$ self-consistency) lifts each
base MLLM's localization accuracy across image, video, and audio, narrowing the
gap to supervised specialists.
}
\label{fig:hook}
\end{figure}

\section{Introduction}
\label{sec:intro}

\begin{figure*}[t]
  \centering
  \includegraphics[width=\textwidth]{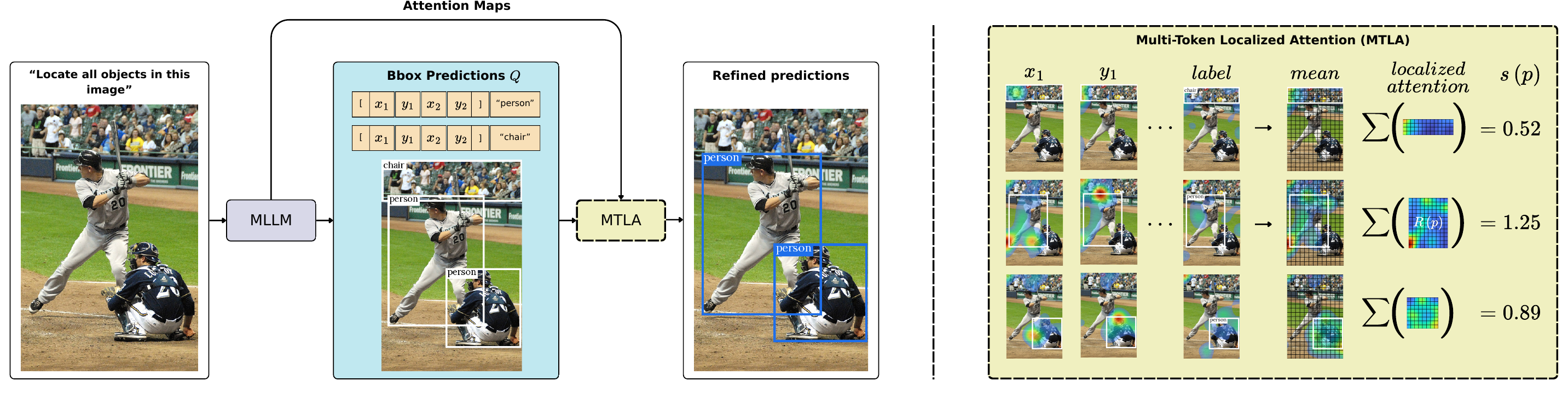}
  \setlength{\abovecaptionskip}{0pt}
  \vspace{0mm}
  \caption{\textbf{Multi-Token Localized Attention (\mtla{}).}
  A pre-trained MLLM localizes objects, emitting per prediction a token
  sequence of box coordinates and a class label. We read the decoder's
  attention from these prediction tokens $Q_p$ onto the input, then
  \emph{restrict} it to the patches inside the model's own proposed region
  $R_p$. The training-free
  score $s(p)$ is the mean over prediction tokens of the attention mass inside
  $R_p$: high when the model attends where it claims (grounded), low when it
  does not (hallucination).}
  \label{fig:teaser}
\end{figure*}

Recent multimodal large language models (MLLMs), including
Qwen~\cite{bai2025qwen25vl},
Gemma~4~\cite{team2025gemma3},
and Audio Flamingo 3~\cite{goel2025af3},
are capable of producing localization outputs, such as spatial
bounding boxes and temporal windows, directly through language generation.
Trained on large-scale grounding corpora such as
RefCOCO~\cite{yu2016refcoco}
and Visual Genome~\cite{krishna2017visualgenome},
these models can localize objects, events, and activities without dedicated
task-specific heads.
The same generation mechanism naturally extends across
modalities, enabling object localization in images, temporal
localization in video, and sound-event localization in audio.

Localization has traditionally been addressed using specialized architectures.
Closed-vocabulary detectors such as
DETR~\cite{carion2020detr}
and open-vocabulary systems such as
GLIP~\cite{li2022glip},
OWL-ViT~\cite{minderer2022owlvit},
and Grounding DINO~\cite{liu2024gdino}
are explicitly designed and optimized for localization tasks. In contrast,
MLLMs provide a unified framework for localization, allowing a single model to
operate across modalities and query types without task-specific architectures or ad hoc training.
As these models continue to improve, an important question emerges: can MLLMs
serve as reliable localization systems?

Despite their growing use for localization tasks, MLLMs are rarely evaluated
using the standard metrics of the detection literature. Existing evaluations
primarily focus on captioning, visual question answering, chat, or
task-specific grounding benchmarks
~\cite{chen2015cococaptions,goyal2017vqav2,liu2024mmbench},
while metrics such as Average Precision (AP), Recall@1, and PSDS1
remain largely unexplored. When evaluated from this perspective, however, a
major limitation becomes apparent. Across the four localization benchmarks
considered in this work, we find that between 58\% and 68\% of the regions
emitted by state-of-the-art MLLMs do not correspond to valid objects or events
in the input (Fig.~\ref{fig:hook}, Table~\ref{tab:coco-models}). This observation highlights a significant reliability gap when
MLLMs are used for localization.

Moreover, practical localization systems require reliable confidence estimates
for individual predictions in order to control the operating point and balance
precision against recall. The same requirement applies when deploying MLLMs
for localization tasks. While the most natural confidence signal is the model's
own token probability, we find that token probabilities exhibit limited
separation between grounded and hallucinated predictions (Fig.~\ref{fig:coco-boxplot}). As a result,
standard confidence estimation techniques provide little guidance for
identifying reliable localization outputs.

Hallucination detection in MLLM outputs has received increasing attention,
particularly in image captioning and related generation tasks. The Summed
Visual Attention Ratio (SVAR)~\cite{jiang2024devils} detects hallucinated
object mentions by aggregating attention from the first sub-token of a
generated object word to the visual tokens. Related approaches study
hallucination detection in text-only language
models~\cite{chuang2024lookback} or modify the decoding process to suppress
hallucinated generations~\cite{leng2024vcd}. More recently, concurrent
work~\cite{nguyen2026beyond} has shown that attention dispersion contains
useful grounding information and trains a patch-level classifier for
hallucination detection. While effective for free-form generation, these
approaches do not fully exploit the localized nature of these outputs, where each prediction is
associated with a specific spatial or temporal region. They either aggregate
attention globally across the entire input, discarding the structure inherent
to localization predictions, or require additional supervision and training.

In this paper, we identify two factors that are essential for reliable
confidence estimation in MLLM localization outputs. First, grounded
predictions attend strongly to modality evidence within their own proposed
region, whereas hallucinated predictions rely more heavily on contextual
distractors elsewhere in the input. This suggests that confidence
should be measured relative to the prediction's own proposed region rather
than globally across the entire input. Second, localization predictions are
inherently multi-token outputs, typically consisting of coordinates together
with a semantic label. Individual prediction tokens often provide noisy or
incomplete evidence, whereas aggregating information across all prediction
tokens yields a substantially more robust grounding signal.

Motivated by these observations, we propose \textbf{Multi-Token Localized
Attention (MTLA)}, a training-free confidence estimation framework for
localization outputs generated by MLLMs. Given a predicted region, MTLA
aggregates decoder attention from all prediction tokens and restricts the
aggregation to modality tokens that fall inside the model's own proposed
region. The resulting score measures the degree to which a prediction is
supported by evidence within the region it claims as justification. MTLA is
architecture-agnostic, requires no retraining, and applies uniformly across
image, video, and audio localization tasks.

Extensive experiments across multiple benchmarks and model families
demonstrate that MTLA provides a highly effective confidence signal for
localization outputs. MTLA consistently improves the separation between
grounded and hallucinated predictions (Fig.~\ref{fig:coco-boxplot}), enables
effective training-free hallucination suppression, and substantially improves
ranking quality under standard detection metrics. The gains hold across modalities and translate
into accuracy: used as a confidence score, MTLA nearly doubles zero-shot COCO
detection AP for an $8$B generalist (from $20.4$ to $37.0$), closing much of the
gap to supervised detectors with no training.

\vspace{4pt}
\noindent\textbf{Our main contributions:}
\begin{itemize}
\item We identify that grounded MLLM predictions attend to modality evidence inside
their own proposed region while hallucinated ones rely on context elsewhere, and
that this signal is spread across all of the generated tokens.
\item We introduce \mtla{}, a training-free score built on this observation
that applies across modalities and model families with no further tuning.
\item \mtla{} outperforms all prior training-free hallucination-detection baselines and, used as a
confidence for re-ranking, substantially improves a generalist $8$B MLLM's zero-shot
localization accuracy, e.g.\ nearly doubling its COCO detection AP ($20.4 \rightarrow 37.0$).
\item We are among the first to benchmark MLLMs as localization systems under
standard detection metrics and benchmarks (AP, Recall@1, PSDS1), and advocate
this protocol as a complement to existing MLLM evaluations.
\end{itemize}

\section{Related Work}
\label{sec:related}

\paragraph{Attention- and representation-based hallucination detection.}
The closest line of work uses internal attention or hidden-state
representations to detect hallucinations in MLLM outputs. The Summed Visual
Attention Ratio (SVAR) of Jiang \etal~\cite{jiang2024devils} scores each
MSCOCO-extracted object word in a generated caption by the total attention
paid to image tokens at a single response position, the first sub-token of
the word's first occurrence, summed over all image-token keys at each
(layer, head). This (layer, head) map can either be aggregated into a single
score or used as a feature map to train a classifier. Lookback Lens~\cite{chuang2024lookback}, uses the ratio of attention paid to the context versus to the model's
own generated tokens, and trains a linear probe that transfers across
models. GLSim~\cite{park2025glsim} takes a different angle, scoring caption
hallucinations from hidden-state cosine similarity between target tokens
and visual tokens, combining global and local
similarity signals. ContextualLens~\cite{phukan2025contextlens} likewise
operates on hidden states, taking the max cosine similarity between an
answer-token embedding and image-patch embeddings at intermediate layers.
These methods aggregate attention or representations over the entire input
and are developed for unstructured outputs (captions, sentences).
Concurrent to our work, Nguyen \etal~\cite{nguyen2026beyond} share our
observation that global aggregation is the weak point, detecting
hallucinations with trained patch-level metrics on captioning and POPE~\cite{li2023pope}. We
differ in restricting attention to the model's own predicted region rather
than measuring full-grid dispersion, in aggregating evidence across all of the
prediction's tokens rather than a single readout, in being training-free rather than
learning a classifier, and in targeting localization tasks scored by AP.

\paragraph{Decoding-time hallucination mitigation.}
A separate family modifies generation to reduce hallucination.
OPERA~\cite{huang2024opera} adds an over-trust penalty during beam search
and uses a retrospection-allocation rollback to escape attention sink
patterns. VCD~\cite{leng2024vcd} contrastively decodes against a
distorted-image branch to suppress language priors.
M3ID~\cite{favero2024m3id} amplifies the mutual-information signal from the
image during sampling. PAI~\cite{liu2024pai} reweights image attention at
inference time, drawing on the observation that image-attention magnitude
is informative about visual grounding. DAMRO~\cite{gong2024damro}
identifies outlier image-token attention patterns associated with
hallucination and suppresses them during generation. These intervene during
generation and require changes to the decoding loop; our method is
post-hoc, scoring predictions after generation completes from the model's own attention maps.

\paragraph{MLLM grounding and localization.}
Grounding MLLMs emit bounding boxes natively as part of their generated
output. Kosmos-2~\cite{peng2023kosmos2} introduces grounded image-text pairs
and trains a multimodal LLM to produce coordinate sequences,
Shikra~\cite{chen2023shikra} adds referential dialogue with spatial inputs
and outputs, and recent MLLM families such as Qwen~\cite{bai2025qwen25vl},
InternVL3.5~\cite{wang2025internvl35}, and Gemma~4~\cite{team2025gemma3}
include grounding data in their training and emit normalized box coordinates directly in their text output. Extending this
to video yields temporal grounding, predicting start/end timestamps for an
event, and moment retrieval, predicting multiple windows. The same generative
interface carries to audio: audio-language models such as Audio Flamingo
3~\cite{goel2025af3} emit temporal spans for sound events from a text prompt.
We treat all of
these uniformly as ``the model emits a region and we ask whether it attended
inside it,'' which is what lets \mtla{} span modalities.

\section{Method}
\label{sec:method}

\begin{figure*}[!t]
  \centering
  \includegraphics[width=\textwidth]{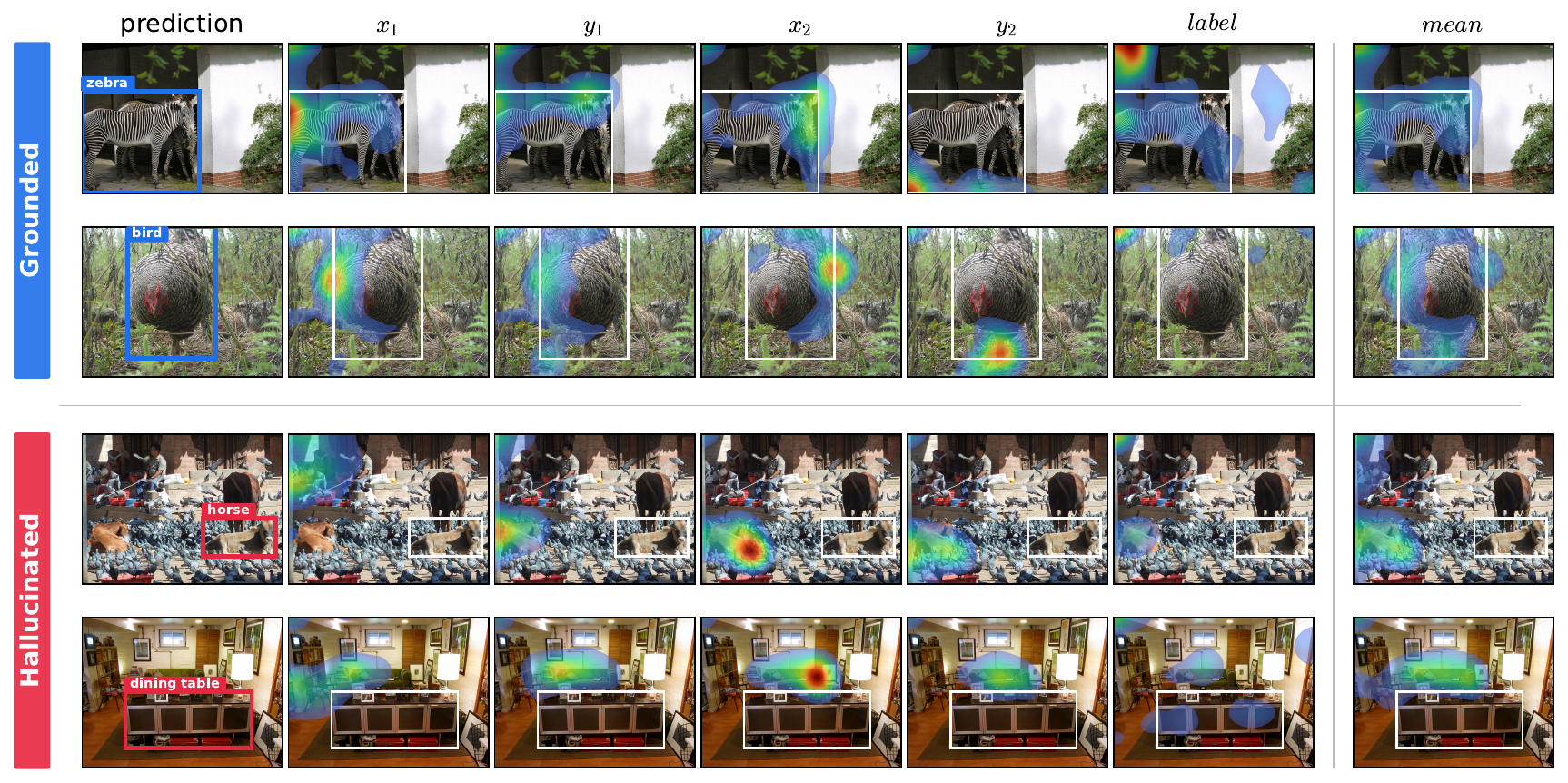}
  \caption{\textbf{Grounding attention is a multi-token signal.}
  Each column shows where the tokens of one of the coordinates ($x_1,y_1,x_2,y_2$) or the label attend; the box marks the proposal
  region $R_p$, and the \emph{mean} column (right of the dashed rule) averages
  them. Any single token gives a partial, noisy view, but jointly the tokens of
  a grounded prediction (\emph{zebra}, \emph{bird}) concentrate their attention
  inside $R_p$, whereas those of a hallucination (a cow labeled ``horse''; a
  phantom ``dining table'') spread it across the scene. See the supplementary material for additional examples.}
  \label{fig:pertoken}
\end{figure*}

In this section, we introduce MTLA, a training-free attention-based confidence score that measures how well a localized prediction is grounded in the input. We first formalize the problem setup and then describe how MTLA uses attention patterns associated with localized predictions to score their grounding confidence (Fig.~\ref{fig:teaser}).

\subsection{Problem setup}
We consider grounding tasks formulated as structured generation problems.
Given an input modality, the MLLM autoregressively generates one or more localized predictions.
Each prediction consists of a region within the input and an associated semantic label.
For example, on open-vocabulary detection the model is asked to enumerate every visible
object as a list of $\{$bounding box, label$\}$ pairs and emits text such as
\begin{verbatim}
[{"bbox_2d": [73, 412, 196, 583],
 "label": "dog"}, ...]
\end{verbatim}
where coordinates are emitted in a $[0,1000]$ space, the convention these grounding MLLMs are trained with. Because their tokenizers represent numbers digit-by-digit, a single coordinate already spans several tokens, which motivates aggregating attention across the prediction's tokens. Rather than force a uniform schema, we let each model answer in the format most natural to it and adapt our method accordingly, so the output stays close to what the model emits unprompted. The exact output syntax differs across model families and modalities; we detail each format and how we parse it in the supplementary material (Sec.~\ref{sec:appendix-formats}).
More generally, each prediction $p$ consists of a region $R_p$ and an associated label.
Depending on the modality, $R_p$ may correspond to a bounding box $[x_1,y_1,x_2,y_2]$ in images or a temporal interval $[t_\text{start},t_\text{end}]$ in video or audio.
We refer to $R_p$ as the prediction's \emph{proposal region}.
A prediction is considered \emph{hallucinated} if its proposal region does not match any ground-truth region according to the task-specific matching criterion (Intersection-over-Union, IoU $\geq 0.5$ throughout).
Our goal is to assign each prediction a scalar score $s(p)$, computed post hoc from the model's own attention, that reflects the likelihood that the prediction is grounded in the input.

\subsection{Multi-Token Localized Attention}
Let the MLLM's transformer comprise $L$ layers and $H$ attention heads.
Let $X={k_1,\ldots,k_N}$ denote the set of modality tokens associated with the input (e.g., visual or audio tokens). 
For each generated response token at position $q$, the transformer produces attention weights $a^{(l,h)}_{q\rightarrow k}$ over tokens $k\in X$ at layer $l$ and head $h$.
We use these attention weights to estimate whether a prediction is grounded in the input.

As a baseline, we define \emph{Global Attention} (\ga{}), the attention signal used by SVAR~\cite{jiang2024devils}, as the attention aggregated over all modality tokens:
\begin{equation}
  \mathrm{GA}^{(l,h)}(q) \;=\; \sum_{k \in X} a^{(l,h)}_{q\rightarrow k}.
  \label{eq:global}
\end{equation}
Our key observation is that grounded predictions attend strongly to modality evidence within their own proposal region. Hallucinated predictions, in contrast, often rely less on evidence inside the proposal region and more on contextual cues distributed throughout the input modality (Fig.~\ref{fig:pertoken}).
Motivated by this observation, we restrict the attention aggregation to modality tokens that fall within the proposal region $R_p$ (Fig.~\ref{fig:teaser}). This yields the \emph{Localized Attention} (LA) signal:
\begin{equation}
\mathrm{LA}^{(l,h)}(q)
=
\sum_{k \in M(R_p)}
a^{(l,h)}_{q\rightarrow k},
\label{eq:la}
\end{equation}
where $M(R_p)$ denotes the set of modality tokens intersecting the proposal region: visual tokens overlapping the predicted box for images, and tokens whose timestamps fall within the predicted span for video and audio. MTLA therefore requires no modality-specific modification beyond constructing this mask (per-modality details in the supplementary material, Sec.~\ref{sec:appendix-formats}).

A grounding prediction is typically represented by multiple output tokens (e.g., bounding-box coordinates and a semantic label in object detection), each of which may capture only part of the underlying evidence. We find that attention patterns associated with a single token may be noisy or incomplete, whereas aggregating attention across all proposal prediction tokens yields a more robust grounding signal (Fig.~\ref{fig:pertoken}).
Let $Q_p$ denote the set of response tokens associated with prediction $p$, i.e., the tokens encoding its region coordinates and, where present, its label. We define \emph{Multi-Token Localized Attention} (MTLA) as the average localized attention across all tokens in $Q_p$:
\begin{equation}
\mathrm{MTLA}^{(l,h)}(p)
=
\frac{1}{|Q_p|}
\sum_{q \in Q_p}
\mathrm{LA}^{(l,h)}(q),
\label{eq:mtla}
\end{equation}

\paragraph{Layer and head reduction.}
To obtain a single confidence score for prediction $p$, we average MTLA across attention heads and over a fixed band of middle transformer layers.
The final prediction score is,
\begin{equation}
s(p)
=
\frac{1}{|\mathcal{L}|}
\sum_{l \in \mathcal{L}}
\frac{1}{H}
\sum_{h=1}^{H}
\mathrm{MTLA}^{(l,h)}(p),
\label{eq:score}
\end{equation}
where $\mathcal{L}$ denotes the selected layer subset.

\paragraph{Self-consistency voting.}
To increase recall, we optionally generate $N$ stochastic rollouts per input, enlarging the candidate pool of predictions, and merge them with non-maximum suppression.
Each kept prediction is scored by the maximum \mtla{} over its cluster, i.e.\ its single highest-scoring rollout. On COCO, where each image yields many predictions, we instead sum the cluster's \mtla{} scores (rewarding regions that recur across rollouts), which improves accuracy over the cluster maximum.
We analyze the effect of the rollout count $N$ in the supplementary material (Sec.~\ref{sec:appendix-rollouts}).

We ablate the proposed components, the proposal-token set, region mask, and layer band, in Sec.~\ref{sec:ablations}.

\section{Experiments}
\label{sec:experiments}
We evaluate \mtla{} across image, video, and audio grounding tasks. We first describe the experimental setup, including benchmarks, models, metrics, and baselines, and then present results showing that it consistently improves hallucination detection and localization quality.
\subsection{Setup}
\label{sec:setup}

We list the benchmarks used in this paper below.

\par\noindent\textbf{COCO Detection.}
COCO val2017 evaluates closed-vocabulary object detection~\cite{lin2014coco}, with $5{,}000$ images and roughly $36{,}000$ ground-truth boxes across $80$ categories.

\par\noindent\textbf{Charades-STA.}
evaluates single-span temporal action grounding in household activity videos~\cite{gao2017tall}. Given a natural-language event description, the model predicts a start/end timestamp pair. The benchmark contains $3{,}720$ queries over $1{,}334$ videos.

\par\noindent\textbf{QVHighlights.}
evaluates multi-segment moment retrieval on $150$-second YouTube clips~\cite{lei2021qvhighlights}. Given a query, the model returns one or more temporal windows. The benchmark contains $1{,}550$ queries over $1{,}519$ videos.

\par\noindent\textbf{AudioSet-Strong.}
evaluates sound-event temporal localization on the strong-label test split~\cite{hershey2021strong,gemmeke2017audioset}, with $14{,}045$ ten-second clips and $116{,}128$ event annotations across $415$ sound classes.

\paragraph{Evaluation protocol and metrics.}
All benchmarks are evaluated in a zero-shot setting, without task-specific fine-tuning or in-context demonstrations. For localization accuracy, we report the standard metric of each benchmark: Average Precision (AP) for COCO detection and QVHighlights, following DETR~\cite{carion2020detr}; Recall@1 for Charades-STA; and PSDS1 for AudioSet-Strong. For hallucination detection, we follow prior work and report \auroc{}. We define a prediction as hallucinated if its proposed region has IoU below $0.5$ with every ground-truth region of the same class.

\paragraph{Models.}
We evaluate \mtla{} on three pre-trained model families: Qwen3-VL-8B-Instruct~\cite{bai2025qwen25vl}, Gemma-4 E4B-it~\cite{team2025gemma3}, and Audio Flamingo 3~\cite{goel2025af3}. The exact architectural configurations and layer-band index choices ($\mathcal{L}$) for each model are detailed in the supplementary material (Sec.~\ref{sec:appendix-impl}).

\paragraph{Baselines.}
Our training-free baselines are post-hoc scores computed from the model's own generated
output, without task-specific training, re-decoding, or auxiliary models (see
the supplementary material, Sec.~\ref{sec:appendix-baselines}, for full
descriptions); supervised specialist detectors are reported separately as references.
As simple confidence baselines, \textbf{raw predictions} assigns every
prediction the same confidence, preserving the model's output order;
\textbf{last\_logp} averages the log-probability of the emitted proposal
tokens; and \textbf{InternalConf}~\cite{jiang2025interpreting} takes the
per-token maximum log-probability across logit-lens layers. The prior
attention- and representation-based detectors we compare against are:
\textbf{\ga{} / SVAR}~\cite{jiang2024devils}, which sums attention from the
first prediction token to all input-modality tokens, without conditioning on
the predicted region; \textbf{GLSim}~\cite{park2025glsim}, which measures
hidden-state similarity between response and visual tokens; and
\textbf{ContextualLens}~\cite{phukan2025contextlens}, which takes the maximum
hidden-state similarity between the target token and visual patches.

\subsection{Image Spatial Grounding on COCO Detection}
\label{sec:coco}

We first evaluate \mtla{} on image-level spatial grounding using COCO val2017~\cite{lin2014coco}. Given an image and the list of 80 COCO categories, the MLLM is prompted to output all detected objects as labeled bounding boxes.
Table~\ref{tab:coco-models} shows the prediction statistics of the original MLLM outputs before any filtering.
Hallucinations are frequent across models; for Qwen3-VL, 68.1\% of 66,590 generated detections do not match any ground-truth object (Fig.~\ref{fig:hook}).

For hallucination detection, we use per-prediction \auroc{}.
As shown in Table~\ref{tab:coco-auroc} and Fig.~\ref{fig:roc}, \mtla{} consistently provides the strongest hallucination signal across both models, improving over SVAR from $82.3$ to $89.0$ on Qwen3-VL and from $67.1$ to $75.2$ on Gemma-4.
Table~\ref{tab:coco-map} shows that hallucination-aware re-ranking directly improves standard COCO detection performance. Starting from the same Qwen3-VL-8B predictions, \mtla{} improves AP from $20.43$ for the raw output order to $32.12$ with a single generation ($N{=}1$), outperforming all post-hoc baselines including SVAR ($28.43$). With self-consistency voting, \mtla{} reaches $36.10$ AP at $N{=}5$ and $37.01$ AP at $N{=}16$, nearly doubling the raw-output AP while remaining training-free. For context we also list zero-shot Gemini API runs~\cite{simedw2024cocogemini} and the supervised DETR-R50 detector ($42.0$ AP)~\cite{carion2020detr}, which use stronger closed models or task-specific training. We ablate the number of self-consistency rollouts $N$ and compare the scaling performance against attention baselines in the supplementary material (Sec.~\ref{sec:appendix-rollouts}).

The Gemini numbers are single-sample ($N{=}1$) API runs, and the DETR baseline is the original supervised DETR-R50 (ResNet-50 backbone) of Carion et al.~\cite{carion2020detr}; both lie outside our zero-shot, training-free, open-model setting and are not head-to-head comparisons.

\begin{figure}[t]
\centering
\includegraphics[width=\linewidth]{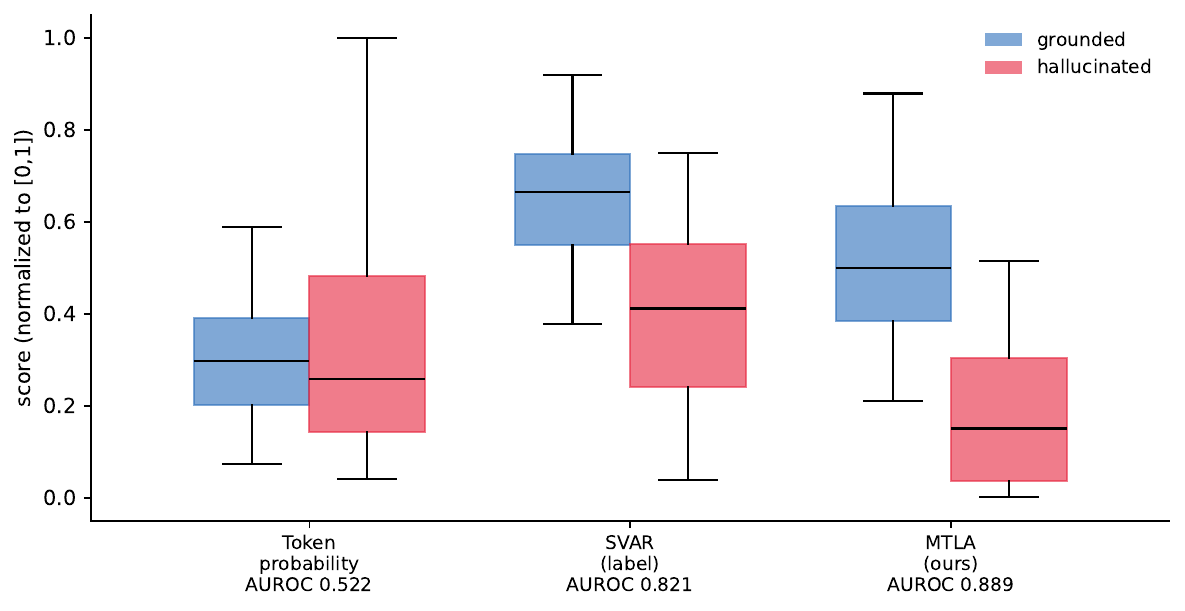}
\caption{Per-prediction score distributions on COCO val2017
(Qwen3-VL-8B), grounded (blue) vs hallucinated (red). Each score is
min-max normalized to $[0,1]$; the box spans the interquartile range and the
bars extend to the $5$--$95$ percentiles. Token probability barely separates the
two classes, SVAR separates them partially, and \mtla{} gives the clearest
separation; see the supplementary material for further analysis.}
\label{fig:coco-boxplot}
\end{figure}

These results highlight a key limitation of token log-probability for grounding: it reaches only $69.0$ \auroc{}, since highly likely coordinate tokens may still describe unsupported objects. This behavior is further illustrated in Fig.~\ref{fig:coco-boxplot}, where grounded and hallucinated predictions have highly overlapping token-probability distributions (per-token breakdown in the supplementary material, Sec.~\ref{sec:appendix-tokenconf}). SVAR improves the separation by using attention, and \mtla{} produces the clearest result.

\begin{table}[t]
\centering
\caption{\textbf{COCO prediction statistics.}
Generated predictions and hallucination rates for each model on COCO val2017.}
\label{tab:coco-models}
\footnotesize
\begin{tabular}{lrc}
\toprule
Model & \#Predictions & Hallu.\ rate \\
\midrule
Qwen3-VL-8B & 66{,}590 & 68.1\% \\
Gemma-4 E4B & 23{,}091 & 37.4\% \\
\bottomrule
\end{tabular}
\end{table}

\begin{table}[t]
\centering
\caption{\textbf{Image hallucination detection.}
Per-prediction \auroc{} on COCO val2017 across different MLLM families.
SVAR is read over each prediction's label tokens (see the supplement,
Sec.~\ref{sec:appendix-formats},~\ref{sec:appendix-extract}).}
\label{tab:coco-auroc}
\footnotesize
\begin{tabular}{lcc}
\toprule
 & Qwen3-VL & Gemma-4 \\
Score & 8B & E4B \\
\midrule
last\_logp & 69.0 & 57.4 \\
InternalConf~\cite{jiang2025interpreting} & 69.0 & 57.4 \\
GLSim~\cite{park2025glsim} & 70.8 & 59.1 \\
ContextualLens~\cite{phukan2025contextlens} & 68.1 & 50.0 \\
SVAR~\cite{jiang2024devils} & 82.3 & 67.1 \\
\textbf{\mtla{} (ours)} & \textbf{89.0} & \textbf{75.2} \\
\bottomrule
\end{tabular}
\end{table}

\begin{table}[t]
\centering
\caption{\textbf{COCO detection accuracy.} AP after re-ranking Qwen3-VL-8B predictions.}
\label{tab:coco-map}
\footnotesize
\setlength{\tabcolsep}{3.5pt}
\begin{tabular}{lcccccc}
\toprule
Score & \mAP{} & \mAP{}$_{50}$ & \mAP{}$_{75}$ & AP$_S$ & AP$_M$ & AP$_L$ \\
\midrule
raw predictions & 20.43 & 32.90 & 20.45 & 5.66 & 24.41 & 47.26 \\
last\_logp & 26.41 & 41.09 & 26.81 & 8.09 & 31.38 & 52.69 \\
InternalConf~\cite{jiang2025interpreting} & 26.41 & 41.09 & 26.81 & 8.10 & 31.38 & 52.69 \\
GLSim~\cite{park2025glsim} & 27.58 & 42.50 & 28.00 & 8.11 & 29.67 & 53.37 \\
SVAR~\cite{jiang2024devils} & 28.43 & 45.52 & 28.82 & 9.81 & 31.05 & 51.05 \\
\textit{Gemini 2.5 Flash}~\cite{simedw2024cocogemini} & \textit{26.1} & \textit{41.7} & — & — & — & — \\
\textit{Gemini 2.5 Pro}~\cite{simedw2024cocogemini} & \textit{34.0} & \textit{51.7} & — & — & — & — \\
\textit{Gemini 3.0 Flash}~\cite{simedw2024cocogemini} & \textit{40.7} & \textit{58.2} & — & — & — & — \\
\midrule
\mtla{} ($N{=}1$) & 32.12 & 48.05 & 32.97 & 11.28 & 34.74 & 55.64 \\
\mtla{} ($N{=}5$) & 36.10 & 56.06 & 36.37 & 13.59 & 39.32 & 60.54 \\
\textbf{\mtla{} ($N{=}16$)} & \textbf{37.01} & \textbf{58.81} & \textbf{36.56} & \textbf{14.48} & \textbf{40.78} & \textbf{61.63} \\
\midrule
\textit{DETR (supervised)}~\cite{carion2020detr} & \textit{42.0} & \textit{62.4} & \textit{44.2} & \textit{20.5} & \textit{45.8} & \textit{61.1} \\
\bottomrule
\end{tabular}
\end{table}

\subsection{Video Temporal Grounding}
\label{sec:video}
We next evaluate \mtla{} on video temporal grounding using Qwen3-VL-8B, on the Charades-STA and QVHighlights benchmarks (Sec.~\ref{sec:setup}). The original MLLM outputs contain substantial hallucination rates: $58.5\%$ on Charades-STA and $66.1\%$ on QVHighlights.

For temporal grounding, we apply the same principle as in images: SVAR aggregates attention over all video frames, while \mtla{} restricts the attention sum to frames inside the model's predicted temporal window.

\begin{figure}[!t]
\centering
\includegraphics[width=\linewidth]{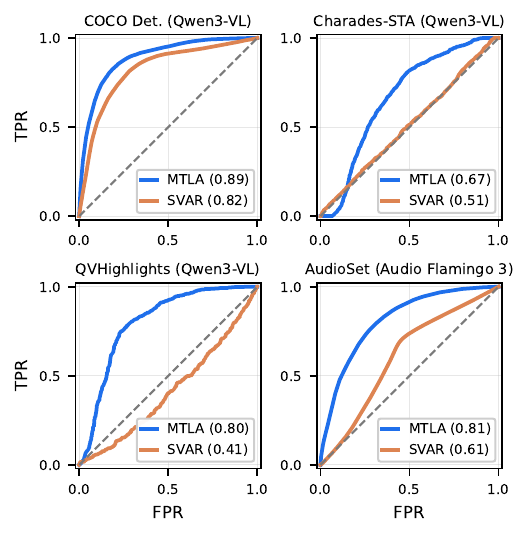}
\caption{\textbf{Hallucination-detection ROC across all four benchmarks.}
\mtla{} (blue) vs.\ \ga{}/SVAR (orange); dashed line is chance, \auroc{} in each
legend. \mtla{} separates grounded from hallucinated predictions more sharply
on every modality.}
\label{fig:roc}
\end{figure}

Figure~\ref{fig:roc} shows that \mtla{} substantially improves hallucination detection on both video benchmarks. On Charades-STA, \mtla{} improves over SVAR from $51.1$ to $66.8$ \auroc{}; on QVHighlights, the gain is even larger, from $41.5$ to $80.0$. This gap indicates that global attention becomes unreliable on longer videos. \mtla{} recovers a more localized and discriminative signal. Token-confidence baselines remain weak ($52.7$--$59.5$ \auroc{}; full numbers in the supplementary material, Table~\ref{tab:video-audio}).

Beyond hallucination detection, \mtla{} can also improve task accuracy by re-ranking the model's stochastic rollouts. We use the same self-consistency recipe as in Sec.~\ref{sec:coco}: top-1 selection for the single-segment Charades-STA task and greedy non-maximum suppression for multi-segment QVHighlights. As shown in Table~\ref{tab:video-grounding}, ranking with \mtla{} improves Charades-STA R@1@0.5 from $44.0$ for a single rollout to $55.4$, and QVHighlights \mAP{} from $24.4$ to $36.6$. In contrast, SVAR yields only limited gains ($43.8$ and $28.1$). Notably, \mtla{} reaches the range of supervised temporal grounding methods, falling between Moment-DETR and QD-DETR on both benchmarks, while remaining zero-shot and training-free. We analyze the interaction between self-consistency and \mtla{} at $N{=}1$ in the supplementary material (Sec.~\ref{sec:appendix-rollouts}).

\begin{table}[t]
\centering
\caption{\textbf{Video temporal grounding accuracy.}
Zero-shot accuracy after re-ranking $N{=}16$ Qwen3-VL-8B rollouts. Charades-STA reports R@1 at IoU thresholds $\{0.3,0.5,0.7\}$; QVHighlights reports \mAP{}. \textit{Italic} denotes supervised methods.}
\label{tab:video-grounding}
\footnotesize
\setlength{\tabcolsep}{4pt}
\begin{tabular}{lccc@{\hskip 10pt}ccc}
\toprule
 & \multicolumn{3}{c}{Charades-STA (R@1)} & \multicolumn{3}{c}{QVHighlights (\mAP{})} \\
\cmidrule(lr){2-4} \cmidrule(lr){5-7}
Score & @.3 & @.5 & @.7 & all & @.5 & @.75 \\
\midrule
raw predictions & 68.0 & 44.0 & 18.8 & 24.4 & 37.0 & 24.4 \\
SVAR~\cite{jiang2024devils} & 68.8 & 43.8 & 18.9 & 28.1 & 43.2 & 27.9 \\
\textbf{\mtla{} (ours)} & \textbf{76.3} & \textbf{55.4} & \textbf{29.4} & \textbf{36.6} & \textbf{54.7} & \textbf{36.6} \\
\midrule
\textit{Moment-DETR}~\cite{lei2021qvhighlights} & \textit{65.8} & \textit{52.1} & \textit{30.6} & \textit{30.7} & \textit{54.8} & \textit{29.4} \\
\textit{QD-DETR}~\cite{moon2023qddetr} & -- & \textit{57.3} & \textit{32.6} & \textit{39.9} & \textit{62.5} & \textit{39.9} \\
\bottomrule
\end{tabular}
\end{table}

\subsection{Audio Temporal Localization}
\label{sec:audio}

We further evaluate \mtla{} on audio temporal localization using AudioSet-Strong~\cite{hershey2021strong,gemmeke2017audioset}. We use Audio Flamingo 3~\cite{goel2025af3} as the audio-capable MLLM, prompted per candidate sound class to emit the temporal windows in which that class is active (full pipeline in the supplementary material, Sec.~\ref{sec:appendix-audio-cascade}). Label matching follows the AudioSet ontology.

The audio results are summarized in Fig.~\ref{fig:roc} (full baseline numbers in the supplementary material, Table~\ref{tab:video-audio}). SVAR's global audio-attention score is substantially weaker ($60.9$ \auroc{}), whereas \mtla{} reaches $81.3$ \auroc{} by restricting attention to frames inside the predicted time span. Beyond ranking, re-ranking and fusing the model's predictions by \mtla{} also improves localization accuracy: as shown in Table~\ref{tab:audio-psds}, \mtla{} improves PSDS1 over the raw output order and over SVAR, approaching a supervised sound-event detector~\cite{schmid2025pretrainedsed} with no task-specific training. For a controlled comparison, we re-evaluate BEATs with its default decoding and no additional post-processing.

\begin{table}[t]
\centering
\caption{\textbf{AudioSet-Strong sound-event localization accuracy.}
Re-ranking and fusing Audio Flamingo 3~\cite{goel2025af3} predictions with different confidence scores.}
\label{tab:audio-psds}
\footnotesize
\setlength{\tabcolsep}{6pt}
\begin{tabular}{lc}
\toprule
Score & PSDS1 \\
\midrule
raw predictions & 0.20 \\
NMS-SVAR & 0.23 \\
\textbf{NMS-MTLA (ours)} & \textbf{0.26} \\
\midrule
\textit{BEATs (supervised)}~\cite{chen2022beats,schmid2025pretrainedsed} & \textit{0.33} \\
\bottomrule
\end{tabular}
\end{table}

\section{Ablations}
\label{sec:ablations}

We ablate the main design choices of \mtla{} to identify where its grounding signal comes from. Unless noted, COCO and video ablations use Qwen3-VL-8B. We present the central ablations here and defer additional ones to the supplementary material.

\subsection{Aggregated Token Set and Region Mask}

We decompose \mtla{} along its two axes, the proposal-token set $Q_p$ and the region mask $M(R_p)$, on COCO Detection and QVHighlights, both with Qwen3-VL-8B (Table~\ref{tab:ablation-decomp}). On QVHighlights, predictions are bare \texttt{[start, end]} timestamps, so the \texttt{label\_mean} and spatial $y_1, y_2$ tokens are undefined ($-$); the two temporal coordinates map to $x_1$ (start) and $x_2$ (end).

\begin{table}[t]
\centering
\caption{\textbf{Aggregated tokens and mask ablation.}
Hallucination-detection \auroc{} for attention-based scores. Rows vary the
proposal-token set $Q_p$; columns compare global attention over the whole
input with attention restricted to the proposal-region mask $M(R_p)$, on COCO (image) and
QVHighlights (video), both with Qwen3-VL-8B. Proposal tokens undefined for the
QVHighlights output format are marked (see text).}
\label{tab:ablation-decomp}
\footnotesize
\setlength{\tabcolsep}{4pt}
\begin{tabular}{lcccc}
\toprule
 & \multicolumn{2}{c}{COCO (image)} & \multicolumn{2}{c}{QVH (video)} \\
\cmidrule(lr){2-3} \cmidrule(lr){4-5}
Proposal tokens & Global & Local & Global & Local \\
\midrule
label\_mean & 81.9 & 83.5 & $-$ & $-$ \\
$x_1$ & 85.4 & 85.4 & 41.5 & 81.4 \\
$y_1$ & 85.3 & 87.0 & $-$ & $-$ \\
$x_2$ & 87.8 & 88.9 & 35.0 & 78.0 \\
$y_2$ & 87.6 & 88.0 & $-$ & $-$ \\
coord\_mean & 87.2 & 88.9 & 37.2 & 80.4 \\
all (ours) & 87.3 & 89.0 & 37.2 & 80.4 \\
\bottomrule
\end{tabular}
\end{table}

The results (Table~\ref{tab:ablation-decomp}) show that both components contribute to the final \mtla{} score. First, replacing global attention with proposal-localized attention consistently improves hallucination detection for every proposal-token set. For example, \texttt{label\_mean} improves from $81.9$ to $83.5$ \auroc{}, \texttt{coord\_mean} improves from $87.2$ to $88.9$, and \texttt{all} improves from $87.3$ to $89.0$.

Second, aggregating across multiple prediction tokens provides a more robust grounding signal than relying on a single generated token. Within the Local column, no individual token (coordinate or label) exceeds $88.9$ \auroc{}, while aggregating all of them (\texttt{all}) reaches $89.0$. The grounding evidence is thus distributed across the prediction's tokens $Q_p$ (Fig.~\ref{fig:pertoken}).

Qwen3-VL emits a label per box, so the label token carries a per-box grounding signal (\texttt{label\_mean} reaches $83.5$ \auroc{} localized) and adding it to the coordinates lets \texttt{all} edge out \texttt{coord\_mean} ($89.0$ vs $88.9$). Models that emit one label for a list of boxes instead share that token across boxes, so it is uninformative per box and \texttt{all} reduces to \texttt{coord\_mean}.

The same pattern becomes even stronger in video. On QVHighlights, restricting attention to the predicted window lifts \auroc{} from $37.2$ to $80.4$ for the \texttt{coord\_mean} proposal tokens, a $+43$ gain, far larger than the image case.

\subsection{Layer Band Selection}

We next examine how the choice of transformer layers affects \mtla{}. 
Following SVAR~\cite{jiang2024devils}, we read attention from L8--21 of $36$, which were identified as where the model integrates and enhances visual 
information, and where grounding-relevant attention is therefore concentrated. 
This single band is fixed once and reused unchanged across every model and modality, rather than tuned per benchmark.

Table~\ref{tab:ablation-layers} compares this band against using the full layer range. 
The band yields a consistent improvement on the video temporal benchmarks ($+5.8$ \auroc{} on Charades, $+4.6$ on QVH) and is essentially neutral on COCO and AudioSet, 
confirming that the grounding signal is strongest in these layers but is not diluted elsewhere. 
Most importantly, \mtla{} does not depend on careful layer tuning: it remains strong with no layer selection at all, the band offering only a small, fully transferable refinement.

\begin{table}[t]
\centering
\caption{\textbf{Layer-band ablation.}
Hallucination-detection \auroc{} for the full layer range vs.\ our middle band,
across all four benchmarks.}
\label{tab:ablation-layers}
\footnotesize
\setlength{\tabcolsep}{4pt}
\begin{tabular}{lcccc}
\toprule
Layer band & COCO & Charades & QVH & AudioSet \\
\midrule
L0--35 (full) & 88.8 & 61.0 & 75.4 & 81.2 \\
L8--21 (ours) & 89.0 & 66.8 & 80.0 & 81.1 \\
\bottomrule
\end{tabular}
\end{table}

\section{Limitations}
\mtla{} requires access to attention maps and a mapping from predicted regions to modality tokens. 
These are not always directly available in optimized inference pipelines.
A further limitation is computational cost. Unlike purpose-built detectors which
localize in a single forward pass, \mtla{} runs a generative MLLM and,
for the self-consistency gains, $N$ stochastic rollouts. This process incurs a substantially higher computational cost in exchange for training-free,
open-vocabulary generality that applies unchanged across different model families and modalities without any tuning.

\section{Conclusion}
\label{sec:conclusion}
We presented \mtla{}, a training-free grounding confidence score for localized MLLM predictions,
built on two observations: grounded predictions attend more strongly to the regions they claim,
and this evidence is distributed across multiple output tokens.
Combining proposal-localized attention with multi-token aggregation,
\mtla{} consistently improves hallucination detection and re-ranking across image, video, and audio grounding.
Used as a re-ranking confidence, it substantially improves a generalist open-source MLLM's zero-shot localization accuracy,
nearly doubling its COCO detection AP and reaching the range of supervised temporal-grounding methods, without task-specific training.

\clearpage
{
    \nolinenumbers
    \small
    \bibliographystyle{ieeenat_fullname}
    \bibliography{main}
}

\clearpage
\appendix
\begin{center}
  {\large\bfseries Supplementary Material}
\end{center}

\section{Implementation Details}
\label{sec:appendix-impl}

\subsection{Models and decoding}
We evaluate three MLLMs. \textbf{Qwen3-VL-8B-Instruct} ($L{=}36$ layers,
$H{=}32$ heads, full attention) is evaluated with the middle-layer band $l \in [8,21]$;
\textbf{Gemma-4 E4B-it} ($L{=}42$, $H{=}8$, hybrid sliding-window attention) is evaluated with the middle-layer band $l \in [8,21]$ (a cross-family/cross-architecture replication);
and \textbf{Audio Flamingo 3}~\cite{goel2025af3} ($L{=}28$, $H{=}28$;
a Qwen2.5-7B language model over an audio encoder) is the audio model, evaluated using all layers ($l \in [0,27]$) as there is no strong visual-like prior for the audio layers. For detection the model is given the image and a prompt listing all
$80$ COCO category names and asked to localize every instance; generation is
capped at $4{,}096$ tokens, allowing up to ${\sim}50$ detections per image
(audio is capped at $512$ tokens). Single predictions use greedy decoding;
the self-consistency rollouts (Sec.~\ref{sec:coco}) sample with temperature
$0.7$ and top-$p\,{=}\,0.95$, one distinct seed per rollout, fused by
vote-weighted non-maximum suppression at IoU $\geq 0.5$ with the combined
confidence $\texttt{vote}\times\mtla{}$. We report $N{=}16$ rollouts as the
headline voting setting.

\subsection{Structured output formats}
\label{sec:appendix-formats}
\mtla{} operates on the model's own generated text, so for each prediction it
must locate the response tokens that encode the region (coordinates) and,
where present, the label. Rather than impose a single rigid schema, we let
each model answer in the format most ``natural'' to it and adapt our parser to
that format; the prompts ask only for the minimal fields the task requires,
so the syntax below stays close to what each model emits unprompted. The model
families emit predictions in different
syntaxes; we recover, per prediction, the token positions of each coordinate
and of the label. Spatial coordinates are emitted in a $[0,1000]$ space (the
training convention of these models, not a runtime rescaling); temporal
coordinates are in seconds. The tokenizers of both text backbones we use
(Qwen3-VL and Gemma-4) split numbers into single-digit
tokens, so a single coordinate value such as \texttt{421} occupies three token
positions; this is why each coordinate contributes several tokens to the
prediction-token set $Q_p$. We note that digit-by-digit tokenization is common
but not universal across LLMs (e.g., some BPE tokenizers group runs of digits),
so the exact number of tokens per coordinate is model-specific.

\paragraph{Image, coordinates-first (Qwen3-VL-8B, Gemma-4).}
A JSON list of objects, with the coordinates emitted \emph{first} and the
label after:

\vspace{6pt}
\begin{minipage}{\linewidth}\footnotesize
\begin{verbatim}
[{"bbox_2d": [x1,y1,x2,y2], "label": "cat"},
 ...]
\end{verbatim}
\end{minipage}
\vspace{6pt}

\noindent Each box carries its own label token, so the label is
per-prediction, attributed to the box in the window up to the next box so a
label is never mis-assigned across objects.

\paragraph{Video (Charades-STA, QVHighlights; Qwen3-VL).}
Numeric temporal spans in seconds with no label: a single
\texttt{[start, end]} pair for single-span Charades-STA, or a list of windows
for multi-segment QVHighlights:

\vspace{6pt}
\begin{minipage}{\linewidth}\footnotesize
\begin{verbatim}
[start, end]                     # Charades
[[s1, e1], [s2, e2], ...]        # QVH
\end{verbatim}
\end{minipage}
\vspace{6pt}

\noindent Models often phrase these loosely (e.g.\
\texttt{27.7 - 29.5 seconds} or \texttt{[1:21, 1:32]}), so we extract the
ordered numeric pairs directly, treating a decimal point or time colon between
digits as part of a single number. With no label token, the \texttt{label\_mean}
tokens are undefined and the two coordinates map to $x_1$ (start) and $x_2$ (end).

\paragraph{Audio (AudioSet-Strong; Audio Flamingo 3).}
The model is queried one sound class at a time and emits the temporal windows in
which that class is active, as numeric \texttt{[start, end]} spans in seconds
(the label is fixed by the query, not generated):

\vspace{6pt}
\begin{minipage}{\linewidth}\footnotesize
\begin{verbatim}
[[s1, e1], [s2, e2], ...]   # queried class
\end{verbatim}
\end{minipage}
\vspace{6pt}

\noindent As in the video case there is no label token, so the
\texttt{label\_mean} tokens are undefined and the two coordinates map to $x_1$
(start) and $x_2$ (end). The candidate classes queried per clip come from the
open-vocabulary cascade in Sec.~\ref{sec:appendix-audio-cascade}.

\subsection{Audio localization pipeline}
\label{sec:appendix-audio-cascade}
Unlike the image and video models, which localize directly from a single
open-vocabulary prompt, Audio Flamingo 3 localizes most reliably when queried
for one sound class at a time. We therefore wrap it in a lightweight
propose-and-localize cascade; \mtla{} is applied unchanged at the final scoring
step. (i)~\textbf{Propose}: we prompt the model to freely list the distinct
sound events it hears, repeated over several stochastic rollouts and unioned, to
recover a high-recall candidate set per clip. (ii)~\textbf{Normalize}: the
free-text candidate labels are mapped to the $456$ canonical AudioSet
class names by an auxiliary language model, in a recall-oriented manner
that keeps all plausible candidates.
(iii)~\textbf{Localize}: for each candidate class we issue a single-class query,
$N{=}16$ stochastic rollouts, and parse the emitted $[\text{start},\text{end}]$
windows. (iv)~\textbf{Score and fuse}: every predicted window is scored by
\mtla{} (and by the baselines) exactly as in the other modalities, and the
rollouts are fused per class by greedy non-maximum suppression at temporal
IoU~$\geq 0.5$. The reported \auroc{} (Fig.~\ref{fig:roc}) treats each predicted
window as one prediction; the PSDS1 in Table~\ref{tab:audio-psds} scores the
fused detections under the standard protocol.

\paragraph{Audio-token time mapping.}
Audio Flamingo 3 encodes audio with a Whisper-style front-end (a $128$-bin
log-mel spectrogram at a $10$\,ms hop) followed by a Transformer audio encoder
and a projector that together emit audio tokens at a fixed rate of $25$\,Hz
($40$\,ms per token). We therefore map the $i$-th audio token to time
$i/25$\,s; this is exact for any clip length (an earlier version assumed every
clip was exactly $10$\,s, which mislocates the inside-region mask on clips
shorter than $10$\,s). Because the encoder is globally attentive, a token at a
window boundary carries some content from just outside it. We accordingly dilate
the inside-region token mask by $\pm 2$ tokens ($\pm 80$\,ms): sweeping the
dilation from $0$ to $\pm 200$\,ms, hallucination-detection \auroc{} is maximized
by the exact (un-dilated) mask while PSDS1 peaks at $\pm 80$\,ms and declines
beyond, so we use $\pm 80$\,ms for the localization-accuracy results and note
that the choice moves PSDS1 by under $0.002$.

\paragraph{Ground-truth label normalization.}
PSDS1 matching is by exact class name. A small number of AudioSet-Strong
ground-truth events are annotated with raw ontology machine identifiers (e.g.\
\texttt{/m/0174k2}) or display-name variants not present in the canonical class
list; we map these to their canonical names before scoring so that correct
predictions are not counted as misses. This GT normalization is applied
identically to our predictions and to the supervised baselines.

\subsection{Extracting localized attention}
\label{sec:appendix-extract}
\mtla{} reads the decoder's self-attention weights, which fused kernels (e.g.\
FlashAttention~\cite{dao2022flashattention}) do not expose, so we run a single \emph{eager}-attention
forward pass over the concatenated prompt and the model's own response,
recording, at each decoder layer, the attention from the prediction's
response-token positions $Q_p$ onto the input-modality tokens (image patches,
or video/audio frames). Given these rows (one $[\text{heads}\times
\text{tokens}]$ matrix per layer per response token), the score is the
\emph{inside-region} attention mass, summed over the masked tokens, averaged
over $Q_p$, meaned over heads, and averaged over the middle layer band:

\vspace{6pt}
\begin{minipage}{\linewidth}\footnotesize
\begin{verbatim}
def mtla_score(attn, mask_idx, band=None):
    # attn:     [n_layer, n_query, n_heads,
    #            n_modality_tok] (1 pred)
    # mask_idx: token ids inside M(R_p)
    # band:     layer indices (default: all)
    # returns:  scalar grounding score s(p)
    x = attn[..., mask_idx]   # keep M(R_p)
    x = x.sum(-1)             # sum over region
    x = x.mean(1)             # mean over Q_p
    x = x.mean(-1)            # mean over heads
    if band is None:          # all layers
        return x.mean()
    return x[band].mean()     # mean over band
\end{verbatim}
\end{minipage}
\vspace{6pt}

The only modality- and model-specific component is the index set
\texttt{mask\_idx} of input tokens that fall inside the proposal $M(R_p)$.

\paragraph{Single-grid encoders (Qwen3-VL, Gemma-4).}
Most of our models place image tokens on a \emph{single} row-major
$\lfloor h/2\rfloor\times\lfloor w/2\rfloor$ grid (merge size $2$ from
\texttt{image\_grid\_thw}), so a box in $[0,1000]$ coordinates maps directly to
the grid cells it overlaps:

\vspace{6pt}
\begin{minipage}{\linewidth}\footnotesize
\begin{verbatim}
def qwen_mask(box, grid_h, grid_w):
    # box:  (x1,y1,x2,y2) in [0,1000]
    # grid: grid_h x grid_w image tokens,
    #       row-major
    # returns: token ids inside the box
    x1, y1, x2, y2 = box
    c0 = int(x1 * grid_w / 1000)
    c1 = ceil(x2 * grid_w / 1000)
    r0 = int(y1 * grid_h / 1000)
    r1 = ceil(y2 * grid_h / 1000)
    return {r*grid_w + c
            for r in range(r0, r1)
            for c in range(c0, c1)}
\end{verbatim}
\end{minipage}
\vspace{6pt}

\noindent The same single-grid logic covers the temporal modalities: for
Qwen3-VL video and Audio Flamingo 3 audio the tokens form a $1$D sequence of frames,
and a frame is inside if its timestamp falls within a predicted window (union
over windows for multi-segment predictions).

\subsection{Baselines}
\label{sec:appendix-baselines}
All baselines are post-hoc scores from the model's own generated output, with
no task-specific training, re-decoding, or auxiliary models, and use the same
proposal-token and layer-band reduction as \mtla{} where applicable.
\begin{description}
\item[raw predictions] assigns every prediction the same score, preserving the
model's emission order (a no-ranking lower bound).
\item[last\_logp] averages, over the proposal tokens, the log-probability the
final output distribution assigns to each emitted token.
\item[InternalConf~\cite{jiang2025interpreting}] takes, per token, the maximum
logit-lens log-probability of the emitted token across layers (a strictly
stronger token-confidence signal).
\item[\ga{}/SVAR~\cite{jiang2024devils}] sums attention from the first response
token to \emph{all} input-modality tokens, i.e.\ the global, mask-free special
case of \mtla{} (read at the first token, footnote~3).
\item[GLSim~\cite{park2025glsim}] combines global hidden-state similarity to the
final prompt state with top-$K$ local similarity to visual tokens, testing
whether the grounding signal lives in attention specifically or in any
model-internal signal.
\item[ContextualLens~\cite{phukan2025contextlens}] takes the maximum hidden-state
cosine similarity between the target token and visual patches, testing whether
the localization effect also benefits hidden-state similarity methods.
\end{description}

\section{Per-Token Probability Detail}
\label{sec:appendix-tokenconf}

Table~\ref{tab:token-conf} gives the per-token-position breakdown referenced
in Sec.~\ref{sec:intro}: the mean probability the model assigns to its own
emitted token, by token position within each prediction component, for
grounded vs.\ hallucinated COCO detections.

\begin{table}[h]
\centering
\footnotesize
\caption{\textbf{Token probability does not separate grounded from
hallucinated predictions.} Mean probability the model assigns to its own
emitted token, by token position within each prediction component, for
grounded vs.\ hallucinated COCO detections (Qwen3-VL-8B, $915$
predictions). The model emits the box coordinates first, then the label.}
\label{tab:token-conf}
\begin{tabular}{lccc}
\toprule
 & \multicolumn{3}{c}{token position (grounded / halluc.)} \\
\cmidrule(lr){2-4}
 & 1st & 2nd & 3rd \\
\midrule
$x_1$ & .66 / .50 & .56 / .37 & .14 / .15 \\
$y_1$ & .87 / .77 & .61 / .48 & .16 / .19 \\
$x_2$ & .91 / .83 & .69 / .54 & .25 / .24 \\
$y_2$ & .88 / .85 & .59 / .54 & .21 / .23 \\
label & 1.0 / .93 & 1.0 / 1.0 & 1.0 / 1.0 \\
\bottomrule
\end{tabular}
\end{table}

The table reveals where the model's uncertainty lives. Qwen3-VL emits the box coordinates first and the label last, so the leading digit of $x_1$ is the very first prediction token and the genuine decision point: its probability is both the lowest ($0.66$) and the most predictive of grounding (grounded $0.66$ vs.\ hallucinated $0.50$). Once $x_1$ is committed, the leading digits of the remaining coordinates ($y_1, x_2, y_2$) follow at higher confidence ($0.87$ to $0.91$) and distinguish grounded from hallucinated boxes far less sharply, since the model has already decided where the object is. Within each coordinate the confidence then decreases from the leading to the trailing digit ($x_1$: $0.66, 0.56, 0.14$). This reflects the coarse-to-fine nature of autoregressive coordinate decoding: the high-order digit fixes the object's approximate position and carries most of the spatial information, while the low-order digits only refine the exact pixel and are inherently less constrained. The class label, emitted after the box, is essentially deterministic ($1.0$ for grounded, $0.93$ for hallucinated) and so carries little separating signal of its own. Taken together, these patterns show that token probability is not a reliable confidence metric for localization: its variation is governed by where a token sits in the decode order and how fine a coordinate it refines, not by whether the prediction is grounded. \mtla{}, in contrast, reads the localization evidence directly.

\section{Region-Size Attention Bias}
\label{sec:appendix-size}

Because attention maps are softmax-normalized, the localized sum $\sum_{k \in M(R_p)} a_{q \rightarrow k}$ is mathematically biased toward larger proposed regions $R_p$ simply because they contain more modality keys. For tiny regions (e.g., small bounding boxes spanning only 1--2 patches), the attention sum is naturally constrained, which partially accounts for the lower single-sample performance on small objects ($\text{AP}_S = 16.9$) compared to large objects ($\text{AP}_L = 57.9$; Table~\ref{tab:coco-map}). We experimented with spatial normalization schemes (e.g., dividing the attention sum by the number of tokens in the mask or by the region area), but found that this degraded overall performance. Normalization amplifies high-frequency attention noise in small regions, leading to false positives. In addition, \mtla{} outperforms the global SVAR baseline across \emph{every} size category by a wide margin (Table~\ref{tab:coco-map}: $\text{AP}_S$ $22.2$ vs.\ $13.8$, $\text{AP}_M$ $46.8$ vs.\ $35.7$, $\text{AP}_L$ $63.5$ vs.\ $53.0$). This holds even on small objects, confirming that \mtla{}'s advantage stems from localized grounding evidence rather than the region-size bias.

\begin{figure*}[tb]
\centering
\includegraphics[width=\textwidth]{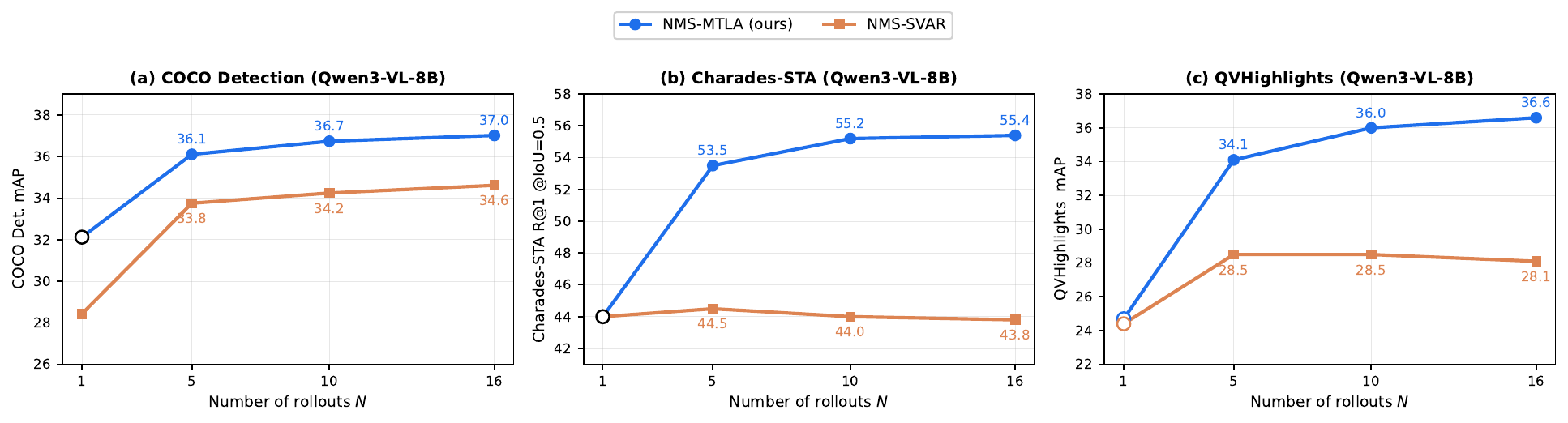}
\caption{\textbf{Self-consistency scaling across modalities.}
(a)~COCO Detection \mAP{} (Qwen3-VL-8B).
(b)~Charades-STA R@1@0.5 and (c)~QVHighlights \mAP{} (Qwen3-VL-8B) as a function of the number of rollouts $N$, comparing MTLA (blue) vs.\ SVAR (orange) selection.}
\label{fig:ablation-scaling}
\end{figure*}

\section{Head Aggregation}
\label{sec:appendix-heads}

Table~\ref{tab:ablation-heads} compares head reduction strategies
(\mtla{}, \texttt{all} proposal tokens, local mask, L8--21).
Both columns use Qwen3-VL-8B.
Mean- and max-over-heads perform within about a point of each other on both
benchmarks (mean is best on COCO, max on Charades-STA), so we use mean throughout.

\begin{table}[h]
\centering
\caption{\auroc{}: head aggregation ablation. Both COCO and
Charades-STA use Qwen3-VL-8B.}
\label{tab:ablation-heads}
\footnotesize
\begin{tabular}{lcc}
\toprule
Head aggregation & COCO & Charades-STA \\
\midrule
mean-over-heads & 89.0 & 66.8 \\
max-over-heads & 87.9 & 67.8 \\
\bottomrule
\end{tabular}
\end{table}

\section{Number of Rollouts and Voting Interaction}
\label{sec:appendix-rollouts}

We investigate the role of self-consistency rollouts $N$ and the voting interaction across both image detection (COCO) and temporal grounding (Charades-STA and QVHighlights) in Figure~\ref{fig:ablation-scaling}.

For COCO detection, Figure~\ref{fig:ablation-scaling}(a) shows detection \mAP{} as a function of the number of stochastic rollouts $N$, comparing \mtla{} and SVAR re-ranking under the same support-weighted fusion, where each kept prediction is scored by the summed \mtla{} (resp.\ SVAR) scores of its cluster. Most of the gain from additional rollouts is realized by $N{=}5$ ($32.1 \rightarrow 36.1$ \mAP{}), after which \mtla{} continues to improve more gradually, reaching $37.0$ \mAP{} at $N{=}16$ (Table~\ref{tab:coco-map}). SVAR benefits from the same enlarged candidate pool but plateaus lower ($28.4 \rightarrow 34.6$ \mAP{}), since its global attention is a weaker per-box grounding signal and cannot separate the true detections from the distractors the larger pool introduces as sharply as \mtla{}.

To understand the interaction between self-consistency rollouts and re-ranking in temporal grounding, we ablate the number of rollouts $N$ and compare \mtla{} to SVAR on the video benchmarks in Figure~\ref{fig:ablation-scaling}(b, c). For the single-segment Charades-STA task, re-ranking is trivial at $N=1$ since only one temporal window is proposed per query; hence, the single-rollout accuracy is equal to the raw predictions ($44.0$ R@1@0.5). For the multi-segment QVHighlights dataset, applying \mtla{} to rank the segments within a single rollout ($N=1$) yields $24.7$ \mAP{} (vs.\ $24.4$ for raw predictions). Pushing to $N=16$ rollouts generates a rich pool of candidates, raising the oracle recall ceiling. Under this setting, the global SVAR baseline fails to isolate the grounded proposals (achieving only $28.1$ \mAP{} on QVHighlights and $43.8$ on Charades-STA), whereas \mtla{} successfully leverages the candidate pool to select grounded segments, reaching $36.6$ \mAP{} and $55.4$ R@1@0.5. This demonstrates that while self-consistency is a key driver of recall, its success is completely dependent on a localized confidence metric like \mtla{} to filter out the generated hallucinations.

\section{Additional Per-Token Attention Examples}
\label{sec:appendix-pertoken}
\begin{figure*}[t]
\centering
\includegraphics[width=\textwidth]{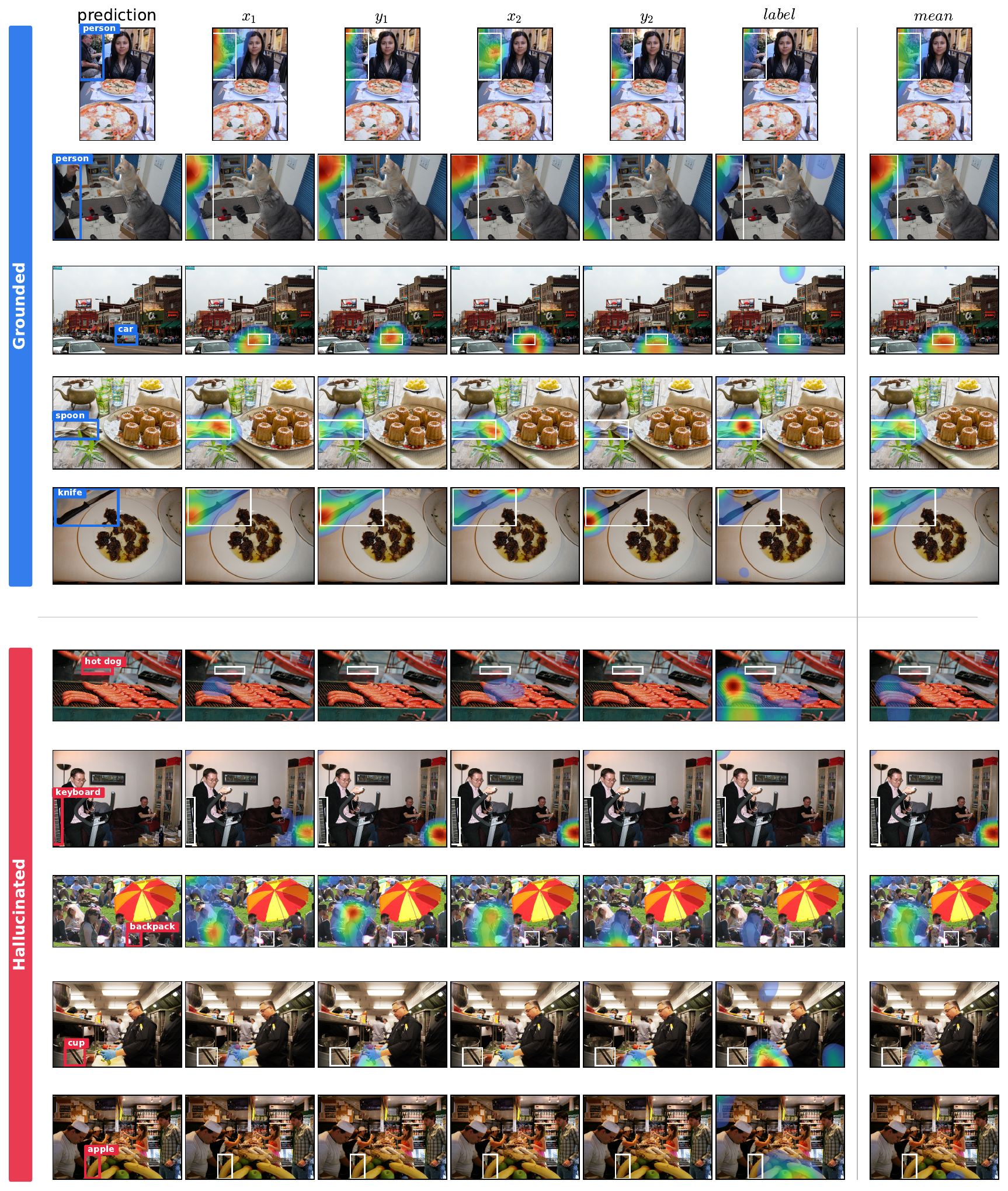}
\caption{\textbf{Per-token attention on additional COCO predictions} (Qwen3-VL,
zero-shot). Top rows are grounded predictions, bottom rows hallucinations;
columns are the coordinate sub-tokens, the label, and their mean. Grounded
predictions concentrate attention inside the proposed box, whereas
hallucinations spread it across the scene.}
\label{fig:pertoken-appendix}
\end{figure*}

Figure~\ref{fig:pertoken-appendix} extends the per-token visualization of
Fig.~\ref{fig:pertoken} to a larger and more diverse set of COCO predictions,
chosen so the proposed region $R_p$ is small enough to make the inside-region
attention pattern clearly visible. For each prediction we show where every
response token attends---the four coordinate sub-tokens
($x_1,y_1,x_2,y_2$) and the label---together with their per-token mean (right of
the dashed rule). For grounded predictions the tokens consistently concentrate
their attention inside $R_p$; for hallucinations the attention instead lands
away from the claimed box, which is precisely the signal \mtla{} reads out.

\section{Token-Confidence Baselines on Video and Audio}
\label{sec:appendix-video-audio-baselines}
Table~\ref{tab:video-audio} reports the full hallucination-detection \auroc{} for
all scores on the video and audio benchmarks, including the token-confidence
baselines (\textbf{last\_logp}, \textbf{InternalConf}) omitted from the
main-paper ROC figure (Fig.~\ref{fig:roc}, which plots \mtla{} vs.\ \ga{}/SVAR).
On both video benchmarks the confidence baselines stay near chance, while
\mtla{} is the strongest score on every benchmark.

\begin{table}[t]
\centering
\caption{\textbf{Temporal hallucination detection.}
\auroc{} on video and audio grounding benchmarks. Charades-STA and QVHighlights use Qwen3-VL-8B; AudioSet-Strong uses the Audio Flamingo 3 cascade (full test set). The token-confidence baselines (\textbf{last\_logp}, \textbf{InternalConf}) are not yet available for the audio cascade (\,---\,).}
\label{tab:video-audio}
\footnotesize
\setlength{\tabcolsep}{4pt}
\begin{tabular}{lccc}
\toprule
 & \multicolumn{2}{c}{Video} & Audio \\
\cmidrule(lr){2-3} \cmidrule(lr){4-4}
Score & Charades-STA & QVHighlights & AudioSet \\
\midrule
last\_logp & 52.7 & 58.1 & --- \\
InternalConf~\cite{jiang2025interpreting} & 52.8 & 59.5 & --- \\
SVAR~\cite{jiang2024devils} & 51.1 & 41.5 & 60.9 \\
\textbf{\mtla{} (ours)} & \textbf{66.8} & \textbf{80.0} & \textbf{81.3} \\
\bottomrule
\end{tabular}
\end{table}

\end{document}